\documentclass[conference]{IEEEtran}
\usepackage{cite}
\usepackage{amsmath,amssymb}
\usepackage{graphicx}
\usepackage{booktabs}
\usepackage{array}
\usepackage{url}

\begin{document}

\title{How Much Fresh Data Should CTR Recalibration Use? A Temporal Evaluation of Post-hoc Calibration}

\author{\IEEEauthorblockN{Rui Liu}
\IEEEauthorblockA{Independent Researcher\\
RL2987@columbia.edu}}

\maketitle

\begin{abstract}
Offline evaluation of click-through rate (CTR) models still defaults to random splits, although deployed models score future traffic and downstream systems consume calibrated probabilities. On Avazu (40.4M impressions) we train logistic regression (LR), DeepFM, and DCN-v2 under two rolling temporal origins and a random split, three seeds per arm, and recalibrate each test day with Platt scaling and isotonic regression fitted on the preceding hours. At origin 1's nearest test day, random splits report LR's expected calibration error (ECE) at 55--62\% of the temporal value in every seed (the deep models' ratio is seed-unstable), and no model's ECE is monotone in temporal distance. With a 6-hour fitting slice, recalibration removes up to 61\% of ECE in badly miscalibrated cells but increases it in about a third of applications (14--17 of 48 across seeds). On fixed evaluation sets, a 12-hour slice cuts harm from 13--17 to 3--5 of 36 applications, and 24 hours raises it again. The 12-hour slice subsampled to the 1-hour slice's size (5\% of the impressions; 20--25$\times$ faster to fit) recovers 88\% of the full slice's gain, so impression count alone does not explain it. A slice-ECE gate shows no benefit out-of-origin, and in a simulated auction recalibration moves profit by only about $-0.21\%$ to $+0.26\%$. Offline protocols should measure calibration temporally, over multiple origins and seeds; in this data, recalibration fitted on a wide time span beat the freshest few hours, which other logs must confirm.
\end{abstract}

\begin{IEEEkeywords}
click-through rate prediction, model calibration, temporal evaluation, post-hoc recalibration, evaluation protocols
\end{IEEEkeywords}

\section{Introduction}

A CTR model in an advertising system outputs not a ranking but a number that other components treat as a probability. Expected-value bidding multiplies it by a value per click; pacing controllers compare predicted against realized click rates to throttle or release budget. In each use the absolute value carries the meaning: a model that ranks well but systematically over- or under-predicts misprices traffic despite a healthy AUC. Calibration---the agreement between predicted probabilities and empirical frequencies---is therefore not a secondary diagnostic for CTR models; it is the property the serving stack consumes.

The offline evaluation practice of the CTR literature is, however, built almost entirely around random train/test splits, with AUC and log loss as headline metrics. Random splits score impressions from the training days, past and future interleaved; a deployed model is trained on the past and applied to the future, under whatever distribution change the intervening days bring. That random splits flatter ranking metrics is partly documented; calibration has received far less attention---yet if a random split reports acceptable calibration while the same model, evaluated temporally, is miscalibrated, the standard protocol certifies the very property it fails to measure.

A second default deserves the same scrutiny. Post-hoc recalibration---fitting a monotone correction (Platt scaling, isotonic regression) on the freshest labeled data and applying it to future predictions---is a widely used recipe. Run \emph{unconditionally}---on every refresh cycle, for every model, whatever the model's current calibration state---it treats recalibration as harmless. It is not: Pan et al.~\cite{pan2020fieldaware} already found Platt scaling and isotonic regression worsening calibration on a date-split Avazu. A correction fitted on a finite slice carries its own error, \emph{added} to the model's predictions when there is little systematic miscalibration to remove---or when the slice's miscalibration points the other way. Whether unconditional recalibration earns its risk is a quality question (its compute, measured in Section~\ref{sec:rq3}, is 1--2~s per cycle at a 12-hour slice), answerable only under a protocol that measures calibration as production experiences it: on future days, one day at a time.

We frame these questions as a measurement study of a recalibration policy's efficiency--quality--cost trade-off, in the sense of benchmarks and protocols for such trade-offs. A recalibration step consumes three resources: fresh labeled data (the fitting slice), compute (one fit--check--apply cycle), and quality risk (the chance the correction makes served probabilities worse); we measure all three in one pipeline on the same data. The slice-length sweep and a size-matched control trace the data axis in two dimensions---how many hours the slice spans and how many labeled impressions it holds---and show that volume alone does not buy quality: a sparse sample spread over 12 hours does nearly as well as the full 1.9M-impression slice, and which hours the slice covers may matter as well. The cycle timings price the compute axis, and the harm counts the risk axis.

Three claims are therefore on the table: that miscalibration grows as the test period moves away from training, as drift accumulates; that recalibration on a recent slice is a safe default remedy; and that, if it is not, a simple gate on a pre-serving measurement can make it safe. This paper measures all three on Avazu with LR, DeepFM, and DCN-v2, under two rolling temporal origins (five-day training windows offset by one day, each evaluated on the next four days, gaps 0 to $+3$) and a random-split arm under the same training cap. For every temporal test bucket we fit Platt scaling and isotonic regression, leak-free, on the six hours immediately preceding it, and carry raw and recalibrated probabilities into a simulated second-price auction. Our research questions:

\begin{itemize}
\item \textbf{RQ1.} How much do random splits inflate discrimination \emph{and} calibration metrics relative to a temporal split?
\item \textbf{RQ2.} As the test day moves farther from the training window, how does calibration error actually behave?
\item \textbf{RQ3.} Does recalibration on the immediately preceding slice recover calibration at each distance---when does it harm instead, how much fresh labeled data should it use, and can a pre-serving gate prevent the harm?
\item \textbf{RQ4.} Do the calibration differences observed in RQ2--RQ3 change realized profit in a simulated auction?
\end{itemize}

In brief: random splits flatter calibration by a day- and seed-dependent amount (RQ1); no model's ECE is monotone in distance (RQ2); recalibration on a 6-hour slice harms about a third of the time, a 12-hour span largely removes the harm, a sparse sample over that span does nearly as well, and slice-ECE gating does not help (RQ3); profit effects range from about $-0.21\%$ to $+0.26\%$ (RQ4). The practical consequence is a change in measurement practice rather than a new method: measure calibration temporally, over more than one origin and seed, and do not recalibrate on short slices by default.

\section{Related Work}

\textbf{Offline evaluation and temporal leakage.} The CTR literature has largely standardized on random splits of benchmark logs, reporting AUC and log loss on held-out impressions from the training period~\cite{zhu2021bars}; recommender-systems work has documented that such splits leak future information and overstate deployed performance~\cite{campos2014time}. Within advertising, McMahan et al.~\cite{mcmahan2013trenches} describe \emph{progressive validation}---prequential, time-ordered evaluation---for tuning production click prediction, and He et al.~\cite{he2014facebook} document model-freshness effects on CTR performance. Temporal-splitting studies for CTR exist (e.g.~\cite{turksoy2024splits}, on Criteo) but report discrimination metrics (AUC, log loss) only. No prior work we know jointly measures, on identical runs, (i) how a random split distorts \emph{calibration} relative to a temporal split, (ii) how calibration error behaves day by day at more than one training origin, and (iii) whether per-period recalibration, refit leak-free on the preceding hours, repairs or harms it---the gap this paper occupies.

\textbf{Post-hoc calibration and calibration methods for CTR.} We evaluate two classical corrections: Platt scaling, a logistic regression on the model's scores~\cite{platt1999}, and isotonic regression, a piecewise-constant monotone map fitted under squared loss~\cite{zadrozny2002transforming}; smoothed isotonic regression interpolates that fit for advertising calibration~\cite{deng2020calibrating}. Niculescu-Mizil and Caruana~\cite{niculescu2005predicting} find isotonic matching or beating Platt once the calibration set exceeds roughly a thousand points in their experiments, though the crossover is model-dependent; our fitting slices, at $\sim$$10^6$ impressions, are far above it. Temperature scaling is a one-parameter relative of the same idea~\cite{guo2017calibration}. CTR-specific method work proposes richer calibration maps: field-aware calibration~\cite{pan2020fieldaware} already evaluates on Avazu split by date (six training, two development, two test days), with calibrators fitted on the development days, and reports Platt scaling and isotonic regression \emph{worsening} field-level calibration there, attributed to shift between development and test days. That recalibration can hurt under temporal shift is therefore not new; we add how often it does so day by day, across two origins and three seeds, how it depends on the amount of fresh data used, and whether a pre-serving gate can prevent it; doubly-adaptive field-level calibration (AdaCalib) learns per-field-value isotonic maps~\cite{wei2022adacalib}. Production accounts describe calibration corrections atop ranking models (a correction layer in~\cite{mcmahan2013trenches}; recalibration after negative downsampling in~\cite{he2014facebook}), evaluated time-ordered; ours is not a new calibrator but a temporal measurement of when the two standard calibrators help, harm, and can be gated.

\textbf{Calibration under distribution shift.} Ovadia et al.~\cite{ovadia2019trust} evaluate predictive uncertainty under dataset shift and find calibration degrades as shift increases, and that post-hoc recalibration fitted on in-distribution validation data does \emph{not} preserve it under shift. That literature typically constructs shift synthetically and fits the correction on source-distribution data; our shift is ordinary day-to-day variation in real ad traffic, and our calibrator is fitted on the freshest pre-serving data.

\section{Experimental Setup}

\subsection{Data}

We use the Avazu CTR dataset~\cite{avazu2014}: 40,428,967 logged impressions spanning ten consecutive days (October 21--30, 2014), with an overall click-through rate of 16.98\%. Each record carries a click label, an hour-resolution timestamp, and categorical fields for site, application, device, and placement; every non-time field is treated as categorical, and an hour-of-day feature is derived from the timestamp. Features are encoded through per-field, frequency-capped vocabularies (at most 20,000 values per field, minimum count 10) fitted on each arm's training rows only---no frequency information from any test period enters the encoding---with unseen values mapped to an unknown token. The same encoded representation feeds all three models in every arm.

\subsection{Evaluation arms}

\textbf{Temporal arms (two origins).} At \emph{origin 1}, models are trained on days 0--4 (Tuesday, October 21 through Saturday, October 25, 2014), with training rows capped at 4,000,000 by deterministic stride sampling: every $k$-th row is kept, with integer $k = \lceil n_{\text{window}}/4{,}000{,}000 \rceil$, which is 2 for both training windows (6.68M and 7.90M rows), so the cap is undershot (3,338,217 rows at origin 1). At \emph{origin 2} the window rolls forward one day: training on days 1--5 (Wednesday, October 22 through Sunday, October 26; 3,948,439 rows after the same cap), testing on days 6--9. At each origin the test buckets are the four days following the training window, indexed by their distance from it: gap 0 through gap $+3$. Buckets at gap $\geq 1$ are evaluated on the full day; the gap-0 bucket is evaluated on hours 6--23 of its day, its first six hours serving as that bucket's recalibration slice (Section~\ref{sec:recal}), giving gap-0 evaluation sets of 3,148,694 (origin 1) and 2,575,774 (origin 2) impressions. Test buckets are otherwise evaluated at full size (no sampling). The tenth day (October 30) is origin 2's gap-$+3$ bucket; origin 1's end at day 8 (October 29). Table~\ref{tab:buckets} lists all eight buckets with their sizes, observed click rates, and signed calibration bias.

\textbf{Random arm.} The same pipeline is trained and evaluated on a random split of the same logs: a reservoir pool split 70/10/20 into train, validation, and test shares, with the training share set to the temporal arms' 4,000,000-row cap (4,000,000 training rows; the cap's stride sampling leaves the temporal arms 3,338,217 rows at origin 1 and 3,948,439 at origin 2, so training size is matched closely only at origin 2); the validation share is never used for model selection or early stopping in this study. The test set contains 1,142,858 impressions, shared across models, interleaving all ten days. This run serves as the RQ1 reference for both origins, against each origin's nearest bucket, gap 0.

\textbf{Seeds and reproducibility.} The headline runs use seed 42. Every arm---origin 1, origin 2, and the random arm---is additionally replicated end-to-end with seeds 43 and 44, giving three training seeds for every cell; we use these to annotate which effects exceed seed noise (Sections~\ref{sec:rq1} and~\ref{sec:rq3}). Every run logs slice statistics; per-impression predictions (for the bootstrap and RQ4) come from seed-42 runs; the sweep (Section~\ref{sec:sweep}) and the size-matched control (Section~\ref{sec:rq3}) are replicated at all three seeds, an aged-window control (Section~\ref{sec:sweep}) uses seed 42, and the origin-1 instrumented rerun reproduced all 36 original metric rows exactly; recomputed profits from the saved predictions reproduce each run's bidding table exactly. Table~\ref{tab:sweep} and Fig.~\ref{fig:f2} report all three seeds; the other tables report seed 42. Code, configurations, and every per-cell result cited here---unrounded metrics for all seeds, slice statistics, gate inputs and decisions, the sweep, the size-matched control, timings, and bootstrap outputs---are released at \url{https://github.com/RuiLiu0129/ctr-recalibration-temporal}.

\begin{table*}[t]
\centering
\caption{Temporal-arm test buckets at both origins. Gap-0 buckets are evaluated on hours 6--23 of their day (their first six hours are the recalibration slice). Signed bias is the raw model's mean prediction minus the observed click rate (negative = under-prediction), for the target bucket and its 6-hour fitting slice. Seed 42.}
\label{tab:buckets}
\scriptsize
\setlength{\tabcolsep}{4pt}
\renewcommand{\arraystretch}{0.72}
\begin{tabular}{lcclrrrrr}
\toprule
 & & & & & & \multicolumn{3}{c}{Signed bias, target / slice} \\
\cmidrule(lr){7-9}
Origin & Bucket & Date (weekday) & Eval.\ rows & Slice rows & Obs.\ CTR & LR & DeepFM & DCN-v2 \\
\midrule
1 & gap 0 & Oct 26 (Sun) & 3,148,694 & 687,198 & 0.1807 & $-0.0044$ / $-0.0033$ & $-0.0033$ / $-0.0050$ & $-0.0029$ / $-0.0033$ \\
1 & gap $+1$ & Oct 27 (Mon) & 3,225,010 & 620,216 & 0.1816 & $-0.0080$ / $-0.0074$ & $-0.0049$ / $-0.0043$ & $-0.0040$ / $-0.0038$ \\
1 & gap $+2$ & Oct 28 (Tue) & 5,287,222 & 743,264 & 0.1522 & $-0.0031$ / $-0.0036$ & $+0.0027$ / $+0.0001$ & $+0.0016$ / $+0.0011$ \\
1 & gap $+3$ & Oct 29 (Wed) & 3,832,608 & 1,007,461 & 0.1566 & $+0.0031$ / $-0.0071$ & $+0.0048$ / $-0.0018$ & $+0.0054$ / $-0.0028$ \\
\midrule
2 & gap 0 & Oct 27 (Mon) & 2,575,774 & 649,236 & 0.1804 & $-0.0091$ / $-0.0071$ & $+0.0027$ / $+0.0044$ & $-0.0017$ / $-0.0008$ \\
2 & gap $+1$ & Oct 28 (Tue) & 5,287,222 & 743,264 & 0.1522 & $-0.0058$ / $-0.0033$ & $+0.0046$ / $+0.0071$ & $+0.0040$ / $+0.0021$ \\
2 & gap $+2$ & Oct 29 (Wed) & 3,832,608 & 1,007,461 & 0.1566 & $+0.0007$ / $-0.0101$ & $+0.0114$ / $+0.0034$ & $+0.0099$ / $+0.0022$ \\
2 & gap $+3$ & Oct 30 (Thu) & 4,218,938 & 659,152 & 0.1693 & $-0.0058$ / $-0.0134$ & $+0.0088$ / $+0.0032$ & $+0.0051$ / $+0.0005$ \\
\bottomrule
\end{tabular}
\end{table*}

\subsection{Models and metrics}

We evaluate three models. \emph{LR} is logistic regression trained with stochastic gradient descent on the log loss over one-hot features (regularization strength set through $C = 1.0$, converted to the SGD penalty $\alpha = 1/(C \cdot n_{\mathrm{train}})$; 3 epochs, batch size 16,384, adaptive learning-rate schedule with $\eta_0 = 0.01$, averaged weights). \emph{DeepFM}~\cite{guo2017deepfm} combines a factorization machine with a deep component; \emph{DCN-v2}~\cite{wang2021dcnv2} is a cross-network architecture (3 cross layers). Both deep models use embedding dimension 16, an MLP of $[128, 64]$, dropout 0.2, 3 epochs, batch size 2,048, Adam with learning rate $10^{-3}$ and weight decay $10^{-5}$. No early stopping on any test period is used, so stopping decisions cannot confound the arm comparison.

We report AUC and log loss as discrimination and overall-accuracy measures, and expected calibration error (ECE)~\cite{naeini2015bayesian} with 15 equal-width bins over predicted probability as the calibration measure---a binned estimator whose bias is documented in~\cite{kumar2019verified,roelofs2022mitigating}; we also report the \emph{signed bias} (mean prediction minus observed rate), which ECE---an unsigned quantity---cannot express. Recalibration is a monotone remap of scores, so its effect concentrates in ECE: across all 48 temporal cells of both origins, Platt and isotonic variants differ from the raw model by at most 0.0002 in AUC, and their signed log-loss deltas span $-0.0009$ to $+0.0006$ (per-cell values in Table~\ref{tab:grid}). AUC and log loss below are therefore raw-model only, and variants are compared primarily on ECE. ECE with equal-width bins is not a proper scoring rule, and the two measures can disagree: in 9 of the 48 applications recalibration moves ECE and log loss in opposite directions. Deltas quoted in the text are computed from unrounded values and may differ in the last digit from differences of tabulated values.

\subsection{Recalibration protocol}
\label{sec:recal}

For each temporal test bucket we fit two recalibrators, Platt scaling and isotonic regression, under a strict leak-free rule: the fitting data is the 6-hour slice immediately preceding the period being evaluated. For buckets at gap $\geq 1$, this is the final six hours of the preceding day. For gap 0, it is the first six hours of the bucket day, and the bucket is then evaluated on the remainder of that day, so no impression is scored by a calibrator that has seen it or anything after it. The slices contain between 620,216 and 1,007,461 impressions (Table~\ref{tab:buckets})---on the order of $10^6$. The recalibrator is refit per bucket, as a production job refreshing on the freshest completed hours would be, facing the same day-to-day shift as the base model. Section~\ref{sec:sweep} varies the slice length from 1 to 24 hours; RQ1--RQ4 use the 6-hour operating point.

\subsection{Bidding simulation}

To test downstream impact (RQ4), each bucket's impressions are replayed through a simulated second-price auction. Every strategy bids its predicted click probability multiplied by a fixed value per click of 1.0. The competing (market) price for each impression is drawn once from a lognormal distribution ($\mu = -2.2$, $\sigma = 0.9$) and shared across all strategies, making comparisons paired impression-by-impression. A strategy wins an impression if its bid exceeds the market price, pays the market price, and earns the value per click if the impression's logged label is a click; profit is total value earned minus total cost. We compare four strategies per bucket: raw model probabilities, Platt-recalibrated, isotonic-recalibrated, and an \emph{oracle} that bids using the ground-truth labels. The oracle is model-independent and unattainable: it bounds the total profit available from \emph{any} prediction improvement in this simulation. Section~\ref{sec:rq4} additionally defines two test-day oracles, both computed from the bucket's own labels and hence undeployable: a \emph{binned calibration oracle}, which remaps raw scores through the bucket's own 15-bin empirical click rates, and an \emph{isotonic test-day oracle}, which fits isotonic regression on the bucket's own impressions---the monotone map minimizing squared error on the test day's labels, fitted and evaluated in-sample.

\section{Findings}

\subsection{RQ1 --- Random splits flatter twice, by a day-dependent amount}
\label{sec:rq1}

Table~\ref{tab:rq1} compares the random arm against origin 1's temporal arm at gap 0, the nearest temporal bucket. The discrimination gap replicates the standard result: random-split AUC exceeds temporal AUC by $+0.0135$ (LR), $+0.0139$ (DeepFM), and $+0.0141$ (DCN-v2), and log loss moves in the same direction (0.3927--0.3988 random against 0.4135--0.4193 temporal).

The calibration gap is proportionally larger, but model-dependent, so we report it per model. At origin 1's gap 0 the temporal arm's ECE is 0.0092 (LR), 0.0069 (DeepFM), and 0.0070 (DCN-v2); the random arm reports 0.0051, 0.0038, and 0.0051---55\%, 55\%, and 74\% of the temporal values, respectively. In seed 42, a random-split protocol would certify two of the three models as roughly twice as well calibrated as the nearest future-day evaluation shows; across seeds this holds robustly only for LR (see below).

Three qualifications scope this result. First, it is measured at one bucket of one origin and is not stable across buckets: across origin 1's four test days, the random arm's AUC advantage ranges from $-0.0042$ to $+0.0157$, and its log-loss advantage \emph{reverses} at gaps $+2$ and $+3$ (the temporal arm's raw log loss is lower there). Much of that reversal is base-rate composition: normalizing log loss by each test set's base-rate entropy removes most of it, though the gap-0 contrast survives (0.8875, 0.8752, 0.8779 temporal against 0.8744, 0.8611, 0.8635 random). Second, gap 0 is the \emph{nearest} bucket, not necessarily the most favorable one: at origin 1 no model's best temporal day is gap 0, and at origin 2 it is the deep models' best day in seed 42 only (Section~\ref{sec:rq2}). Third---and most consequential---origin 2 shrinks the nearest-day contrast. At origin 2's gap 0 the AUC differences (random minus temporal) are $-0.0002$, $+0.0008$, and $-0.0009$, and the random split reports 52\%, 96\%, and 117\% of the temporal ECE (Table~\ref{tab:rq1} caption), so for LR the ECE flattery is as large as at origin 1. Training volume is a partial confound: the random arm trains on 20\% more rows than origin 1 (4.00M against 3.34M) but only 1\% more than origin 2 (3.95M), so part of origin 1's gap-0 AUC contrast may reflect training size; at the nearly size-matched origin 2 the gap-0 AUC contrast vanishes, while LR's ECE is still understated by half and gaps $+2$/$+3$ keep AUC advantages of $+0.0087$ to $+0.0189$ (seed 42). Away from gap 0 the picture is more uniform than the nearest-day comparison suggests: across all 24 temporal cells of both origins, the random split reports lower ECE in 22 in seed 42 (the exceptions are DCN-v2 at origin 1, gap $+2$, and origin 2, gap 0) and in 20 and 23 in seeds 43 and 44 (same-seed arms compared), and at origin 2's gaps $+2$/$+3$ its AUC advantage is $+0.0087$ to $+0.0189$, larger than at origin 1's gap 0. Across the random arm's three seeds, the origin-1 gap-0 AUC inflation is stable ($+0.0135$ to $+0.0163$ across models and seeds). The gap-0 ECE ratio is seed-stable only for LR (0.55/0.59/0.62 across seeds); for the deep models it swings (DeepFM 0.55/0.61/0.89; DCN-v2 0.74/1.04/0.58---at seed 43 the random arm's DCN-v2 ECE \emph{exceeds} the temporal arm's); we rest no claim on the deep models' ratio. How much the random split flatters depends on which days, and which training window, are involved, not on a constant of the protocol.

\begin{table}[t]
\centering
\caption{Random split vs.\ temporal split at gap 0, origin 1 (raw model). Origin 2's gap-0 temporal values are LR 0.7457/0.4122/0.0097, DeepFM 0.7562/0.4059/0.0040, DCN-v2 0.7558/0.4064/0.0044 (AUC/log loss/ECE). Seed 42.}
\label{tab:rq1}
\footnotesize
\begin{tabular}{llrrr}
\toprule
Model & Arm & AUC & Log loss & ECE \\
\midrule
LR & Random & 0.7455 & 0.3988 & 0.0051 \\
LR & Temporal (gap 0) & 0.7320 & 0.4193 & 0.0092 \\
DeepFM & Random & 0.7570 & 0.3927 & 0.0038 \\
DeepFM & Temporal (gap 0) & 0.7431 & 0.4135 & 0.0069 \\
DCN-v2 & Random & 0.7549 & 0.3938 & 0.0051 \\
DCN-v2 & Temporal (gap 0) & 0.7408 & 0.4148 & 0.0070 \\
\bottomrule
\end{tabular}
\end{table}

\subsection{RQ2 --- Calibration error is not distance-driven, nor fixed by the day}
\label{sec:rq2}

\begin{figure}[t]
\centering
\includegraphics[width=0.8\columnwidth]{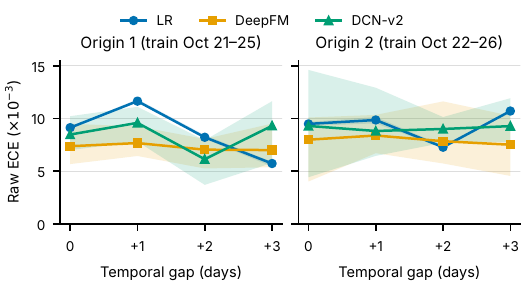}
\caption{Raw ECE by temporal gap at two training origins: mean of seeds 42/43/44 (lines) and min--max range (bands). LR is seed-stable and non-monotone at both origins; the deep models' bands overlap across gaps and their seed-42 peaks do not reproduce.}
\label{fig:f2}
\end{figure}

We expected ECE to grow with the gap between training and test periods, as drift accumulates. Figure~\ref{fig:f2} (three-seed means with min--max bands) and the raw column of Table~\ref{tab:grid} show that this hypothesis was \textbf{not supported} for any model at either origin. LR, whose raw-ECE seed standard deviation is at most 0.0003, gives the clearest picture: at origin 1 it peaks at gap $+1$ and then falls (seed means 0.0091, 0.0116, 0.0082, 0.0057); at origin 2 it dips at gap $+2$ and peaks at gap $+3$ (0.0095, 0.0099, 0.0073, 0.0107). The deep models' curves do not reproduce across seeds. Their seed-42 peaks (gap $+1$ at origin 1, gap $+2$ at origin 2, with origin 2's minimum at gap 0) do not recur: in seed 43 both peak at gap $+3$ at origin 1, and in seeds 43 and 44 origin 2's gap 0 is among their worst buckets. On seed means the deep models are nearly flat at origin 2 (DeepFM 0.0075--0.0084, DCN-v2 0.0088--0.0093), with seed standard deviations up to 0.0031 at origin 1 and 0.0051 at origin 2. Averaged across models and seeds, raw ECE peaks at gap $+1$ at origin 1 (0.0083, 0.0096, 0.0071, 0.0073) and is flat at origin 2 (0.0089, 0.0090, 0.0080, 0.0092).

With a single origin, distance and day identity are perfectly confounded; a second origin one day later gives a second, heavily overlapping view rather than a separation. Three calendar days are evaluated at both origins. October 28 and 29 are full days at both origins, and LR calibrates each differently by origin (seed-mean raw ECE, origin 1 $\rightarrow$ origin 2): 0.0082 $\rightarrow$ 0.0099 and 0.0057 $\rightarrow$ 0.0073, ratios of 1.20 and 1.27, an order of magnitude above LR's seed variation. (October 27 also differs, 0.0116 $\rightarrow$ 0.0095, but origin 2 evaluates only its hours 6--23.) For LR, moving the training window by one day changes a fixed day's calibration about as much as moving between days does. Two heavily overlapping origins cannot separate distance, date, weekday, and training-window membership; they show only that neither distance nor date alone describes the measured trajectories. The deep models' seed-42 differences are larger (up to $1.9\times$, on October 29), but on seed means they shrink to 1.0--1.4$\times$ and cannot be separated from seed variation.

The signed bias (Table~\ref{tab:buckets}) tells the same story in directional form. At origin 1, all three models \emph{under}-predict on the gap-0 (Sunday) and gap-$+1$ (Monday) buckets, are near zero or mixed at gap $+2$, and \emph{over}-predict at gap $+3$: the bias flips sign across days. At origin 2 (seed 42) there is no shared flip: the deep models over-predict on nearly every bucket (up to DeepFM $+0.0114$ at gap $+2$), while LR under-predicts on three of four. Neither origin's pattern is a steadily growing one-directional drift, and a fixed day's bias moves with the origin: on October 29 LR's over-prediction shrinks from $+0.0031$ to $+0.0007$.

One competing explanation can be checked, though not cleanly. At both origins the gap-0 and gap-$+1$ weekdays are absent from the training window, which could explain origin 1's gap-$+1$ peak; yet at origin 2 the seed-stable LR peaks on Thursday and is lowest on Wednesday, both weekdays its window contains. With ten days of data weekday-absence coincides exactly with ``gap $\in \{0, +1\}$'' (as a harm predictor it scores 0.333 pooled), so this checks one binary predictor and cannot separate weekday composition from distance or date. The two origins thus argue against planning on monotone drift, without refuting the decay-with-delay pattern reported in production~\cite{he2014facebook}.

\subsection{RQ3 --- Recalibration helps conditionally, sometimes harms, and resists gating}
\label{sec:rq3}

Table~\ref{tab:grid} gives the full grid at both origins---raw, Platt, and isotonic ECE for every model and bucket: 48 calibrator applications in total, 24 per origin.

\begin{figure}[t]
\centering
\includegraphics[width=0.88\columnwidth]{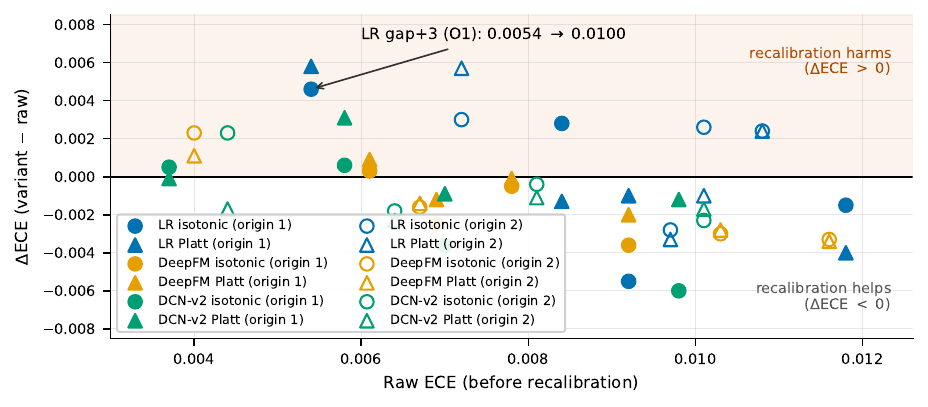}
\caption{Recalibration outcome vs.\ initial miscalibration, both origins (filled markers: origin 1; open markers: origin 2). Harm occurs at low, moderate, and high raw ECE (up to 0.0108) and follows no single threshold across origins. Seed 42, 6-hour slice.}
\label{fig:f3}
\end{figure}

\emph{Where the raw model is miscalibrated, recent-slice recalibration usually works.} At origin 1, isotonic produces large reductions in five cells, all with raw ECE $\geq 0.0069$: DCN-v2 at gap $+1$ ($-61\%$, 0.0098 $\rightarrow$ 0.0038), LR at gap 0 ($-60\%$, 0.0092 $\rightarrow$ 0.0037), DCN-v2 at gap 0 ($-51\%$), DeepFM at gap $+1$ ($-39\%$), and DeepFM at gap 0 ($-36\%$). At origin 2 the same pattern holds at the deep models' worst buckets (gap $+2$, raw ECE 0.0116 and 0.0101), where both calibrators cut error to 0.0078--0.0084 (seed 42; these cells are within seed noise). But outcomes at high raw ECE still span the full range (Fig.~\ref{fig:f3}): at origin 1, isotonic gains only $-13\%$ for LR at gap $+1$ and $-6\%$ for DeepFM at gap $+2$, and it \emph{harms} LR at gap $+2$ (0.0084 $\rightarrow$ 0.0112). Large gains are a property of some cells, not of the method.

\begin{figure}[t]
\centering
\includegraphics[width=0.92\columnwidth]{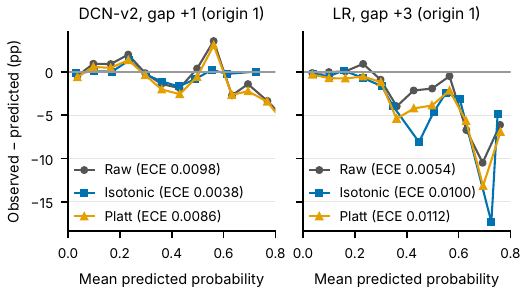}
\caption{Calibration gap (observed minus predicted click rate, percentage points) per 15-bin before and after recalibration, in the grid's best isotonic cell (left) and LR's worst cell (right); origin 1, seed 42, bins with $\geq$1,000 impressions. Recalibration pulls gaps toward zero in the first case, away in the second.}
\label{fig:f4}
\end{figure}

\emph{Where slice and target disagree, the same recipe adds error.} In seed 42, eight of the 24 applications increase ECE at each origin---16 of 48 in total; seeds 43 and 44 give 14 and 17 of 48, so the pooled rate of about a third is seed-stable, but its split between origins is not (3 and 4 harms at origin 1, 11 and 13 at origin 2). In seed 42 the harm concentrates at origin 1's gap $+3$, where \emph{all six} applications harm; in seeds 43 and 44 only LR's two do. The worst cell in either grid is LR at gap $+3$ under Platt, where ECE more than doubles, 0.0054 $\rightarrow$ 0.0112 ($+107\%$), with isotonic in the same cell at 0.0100 ($+85\%$); Fig.~\ref{fig:f4} contrasts its per-bin calibration gaps with those of the grid's best isotonic cell. At origin 2 (seed 42), gap $+3$ harms only LR (both calibrators, 0.0108 $\rightarrow$ 0.0132); the remaining harms are LR at gap $+1$ (isotonic) and gap $+2$ (both calibrators)---five of LR's eight origin-2 applications harm in all---and the deep models at gap 0, whose raw ECE is already the origin's lowest (0.0040--0.0044), pushed up in three of four applications. Outside gap $+3$, Platt helps in all nine of its origin-1 cells, including a 34\% cut at LR gap $+1$ (0.0118 $\rightarrow$ 0.0078) where isotonic manages 0.0103.

\emph{Seed stability: which of these cells are real?} LR is seed-stable at both origins (per-cell ECE standard deviation at most 0.0003; every LR recalibration sign is consistent across seeds, and origin-2 LR harms at gaps $+1$ to $+3$ in every seed). The deep models are not: raw-ECE seed standard deviations run 0.0014--0.0031 at origin 1, and by the criterion ``seed-42 sign flips across seeds, or $|\Delta\mathrm{ECE}|$ below two standard deviations,'' 11 of 24 origin-1 cells and 14 of 24 origin-2 cells lie within seed noise, all of them deep-model cells. The headline cells are all seed-stable: LR gap $+3$ harm (Platt $+0.0058$, isotonic $+0.0045$), LR gap $+2$ isotonic harm ($+0.0028$), and the gap-$+1$ isotonic gains for DCN-v2 ($-0.0060$) and DeepFM ($-0.0037$). We rest RQ3's conclusions on the LR grid, the large origin-1 deep-model gains, and the pooled harm rate (14--17 of 48 across seeds), not on small deep-model cells or on the split between origins.

\emph{Uncertainty on $\Delta$ECE itself.} Applying the RQ4 paired block bootstrap (200 blocks, 10,000 resamples) at both origins, 43 of the 48 $\Delta$ECE intervals exclude zero, including every seed-stable headline cell (e.g.\ LR gap $+3$ under Platt, $+0.0058$, 95\% CI $[+0.0043, +0.0066]$). The five intervals that include zero are all among the 25 cells the seed analysis flags as noise, but the bootstrap calls the other 20 significant: these intervals hold the trained model and fitted calibrator fixed, so with 2.6--5.3 million impressions per bucket nearly every difference is significant. Seed replication, not test-set sampling, is the binding uncertainty here.

\emph{Why harm happens: the arithmetic points away from sample size.} The 6-hour slices hold 0.62--1.01 million impressions: Platt's two parameters are then estimated almost exactly, yet Platt doubled LR's gap-$+3$ ECE, and the slices are three orders of magnitude above the size at which isotonic matches Platt in~\cite{niculescu2005predicting}. Starvation is thus an unlikely sole cause, though estimation error in sparse score regions is not ruled out. A systematic account fits better: the correction is fitted on one slice (for gap $\geq 1$, the previous day's final six hours---an evening slice) and applied to a different target (a full new day, with different hour composition, traffic mix, and click rate)---a mechanism testable directly because the instrumented runs logged each slice's signed bias (Table~\ref{tab:buckets}). At origin 1's gap $+3$, the slices' biases are negative for all three models ($-0.0071$, $-0.0018$, $-0.0028$ for LR, DeepFM, DCN-v2) while the target day's are positive ($+0.0031$, $+0.0048$, $+0.0054$): the slice-fitted map shifts predictions upward on a day that already over-predicts. Two parameter-free rules make this precise over the 48 seed-42 applications. ``Harm iff the slice's and target's biases differ in sign'' scores 0.833 (0.917 at origin 1, 0.750 at origin 2). ``Harm iff removing the slice's bias from the target day would leave a larger absolute bias'' scores 0.875 (0.917 and 0.833), against 0.667 for ``never harm''; all 10 applications it flags are harmed. Its six misses are small-bias isotonic cells and origin 2's low-ECE deep-model gap-0 cells. Both rules need the target day's bias, so neither is deployable; they show what a pre-serving signal must anticipate. The second rule is only partly seed-stable: with each seed's own slice statistics it classifies 46 of 48 applications in seed 43 (against 34 for ``never harm'') but only 33 in seed 44 (against 31), and at origin 1 in seed 44 it falls below the trivial rule (16 against 20 of 24). Isotonic's greater flexibility can express more of the mismatch---consistent with the LR--isotonic exceptions at gaps $+1$ and $+2$, where the parametric calibrator is safer.

\emph{Span, not size: a size-matched control.} For each fixed bucket (gaps $+1$ to $+3$), five uniform subsamples of its 12-hour slice were drawn down to exactly the 1-hour slice's impression count (median 93K, 5\% of the 12-hour volume), holding size at the 1-hour level while keeping the 12-hour span; run at all three seeds, at seed 42 mean $\Delta$ECE is $-0.0024$ for the full 12-hour slice, $-0.0001$ for the true 1-hour slice, and $-0.0021$ for the size-matched subsamples, which recover 87\% of the full slice's gain over the 1-hour slice, $(\Delta_{\mathrm{sub}}-\Delta_{1\mathrm{h}})/(\Delta_{12\mathrm{h}}-\Delta_{1\mathrm{h}})$ (0.80--0.99 across origin$\times$calibrator), and match its harm count (4 of 36, against 15 for the true 1-hour slice); adding the gap-0 cells, whose 1-hour slice is the first hour of the test day itself, gives 88\%. Across seeds, recovery is 88\% pooled (87--88\% per seed; 0.72--1.00 across seed$\times$origin$\times$calibrator), and subsample harms (4, 7, 4 of 36 by seed) track the full slice's (4, 5, 3), against 15--18 for the true 1-hour slice. In this control, impression count alone does not explain the improvement; time span does more. Span changes several things at once---diurnal coverage, the number of distinct hours sampled, and the data's average age; one reading is that hour-to-hour shifts make the number of distinct hours, not of impressions, the effective sample size for the day's bias. We have not separated these.

\emph{Predicting harm: simple rules degrade out-of-origin.} Three one-variable harm rules fitted on origin 1 and applied frozen to origin 2 all lose accuracy: ``harm iff raw ECE $\leq 0.0065$'' (the target day's ECE, known only after serving) falls from 0.917 to 0.667, exactly the ``never harm'' baseline; ``harm iff gap $= +3$'' falls from 0.917 to 0.583, below it; and ``harm iff the bucket's weekday is absent from training'' (equivalent here to gap $\in\{0,+1\}$) scores 0.333 pooled.

\emph{The deployable gate shows no benefit out-of-origin.} A production gate must threshold a quantity available \emph{before} serving; the natural candidate is the raw model's ECE on the fitting slice. Fitted on origin 1 ($\tau_{\text{slice}} = 0.0061$), the gate essentially matches always applying isotonic there (mean ECE 0.0063 against 0.0064; never recalibrating, 0.0076). Frozen and applied to origin 2, it gives mean ECE 0.0082 against 0.0081 for always-isotonic and 0.0083 for never---no detectable benefit---while always-Platt reaches 0.0075, although Platt was the weaker calibrator at origin 1 (0.0074 against 0.0064): \emph{which} calibrator to apply flips between origins. As a harm classifier on isotonic applications the gate falls from 0.750 to 0.500, below the 0.583 of ``never harm.'' This test is out-of-origin, not out-of-time: origin 2's slices for gaps $+1$ to $+3$ are origin 1's for gaps $+2$ and $+3$. The failure has a measurable cause: slice ECE correlates with target-day ECE at only $r \approx 0.39$ pooled (0.342 and 0.379 per origin; up to 0.517 across the other seeds; with 12 model--buckets, 95\% CI roughly $-0.29$ to $0.77$). Gates on the slice's signed bias (its magnitude, either direction alone, and slice ECE minus $|$bias$|$), fitted and frozen the same way, choose between raw and isotonic, so their baseline is always-isotonic: two of the four beat it at both origins (positive-bias gate 0.0061 and 0.0079, firing on 9--10 of 12 buckets; ECE-minus-$|$bias$|$ gate 0.0060 and 0.0078), by 0.0002--0.0004, within seed noise, and none beats always-Platt. Reliability-curve shape and cross-slice agreement remain untested as gates. In policy terms, the slice-ECE gate is dominated: it spends a measurement and a threshold decision every cycle yet does no better than the fixed always-isotonic policy and worse than always-Platt. Adaptivity is itself an efficiency claim, and here it does not pay for its decision.

\emph{What this costs: the compute axis.} Table~\ref{tab:cost} prices both stages of the pipeline on our hardware---a 2-vCPU virtual machine on an AMD EPYC 9D25 host (Python 3.12.3, scikit-learn 1.9.1, PyTorch 2.14.1+cpu, NumPy 2.5.4, SciPy 1.18.1). Training an arm end to end costs 1,291--1,457~s; a recalibration cycle (fit plus application) costs 1.1--1.9~s at a 12-hour slice and 2.3--3.3~s at 24 hours. The same fit's timing varies by 23--52\% between runs, so speed-up ratios are computed within runs. Refit cost grows roughly linearly in slice volume: a 12-hour fit costs a median 1.41~s (Platt) or 0.87~s (isotonic), against 0.09/0.03~s at 1 hour, and applying a fitted calibrator adds 0.2--0.5~s per bucket at any slice length. At a hypothetical hourly cadence for three models (2,160 cycles per month; our protocol refits daily), even the 24-hour operating point, one calibrator per cycle, totals only about two CPU-hours per month. Both the harm and the failed remedy are signal problems, not compute problems.

\begin{table}[t]
\centering
\caption{Measured compute on the 2-vCPU VM. Training: wall-clock per model (seed-42 run logs; the full arm including evaluation totals 1,291~s at origin 1 and 1,457~s at origin 2). Refit: median fit time per bucket with median slice volume; 1/3/12/24 h from the seeds-43/44 sweep instrumentation (48 fits per cell, 36 at 24 h). $^\dagger$From an earlier timing probe on two origin-1 slices (median of seven), not directly comparable. $^\ddagger$Size-matched control (Section~\ref{sec:rq3}), seed 42, gaps $+1$ to $+3$.}
\label{tab:cost}
\scriptsize
\setlength{\tabcolsep}{3pt}
\renewcommand{\arraystretch}{0.82}
\begin{tabular}{llr}
\toprule
Item & Detail & Time (s) \\
\midrule
Training, origin 1 & LR / DeepFM / DCN-v2 & 10.5 / 205.7 / 599.8 \\
Training, origin 2 & LR / DeepFM / DCN-v2 & 13.1 / 257.6 / 703.0 \\
\midrule
Refit, 1 h (87K impr.) & Platt / iso. & 0.085 / 0.032 \\
Refit, 3 h (275K) & Platt / iso. & 0.26 / 0.11 \\
Refit, 6 h (0.69--1.01M)$^\dagger$ & Platt / iso. & 0.93--1.12 / 0.27--0.63 \\
Refit, 12 h (1.86M) & Platt / iso. & 1.41 / 0.87 \\
Refit, 24 h (3.83M) & Platt / iso. & 2.80 / 2.11 \\
12 h, control runs (1.86M)$^\ddagger$ & Platt / iso. & 1.14 / 0.58 \\
12-h span, 1-h size (93K)$^\ddagger$ & Platt / iso. & 0.057 / 0.023 \\
\bottomrule
\end{tabular}
\end{table}

\subsection{The data-efficiency frontier: harm drops sharply at a 12-hour slice}
\label{sec:sweep}

The 6-hour slice of the main protocol is a choice, not a constant of the problem. We refit both calibrators on slices of 1, 3, 6, 12, and 24 hours immediately preceding each bucket (all three seeds, both origins; 6 hours reproduces RQ3). The gap-0 bucket must be treated separately: its slice is the first hours of the bucket day, so a longer slice shrinks its evaluation set (to hours 12--23 at 12 hours; none at 24) and changes its raw ECE (origin-2 LR 0.0097 at 6 hours, 0.0076 at 12). Table~\ref{tab:sweep} therefore reports gaps $+1$ to $+3$ (fixed evaluation sets), with the median number of fresh labeled impressions each refit consumes.

On these cells there is no gradual dose--response, in any seed. From 1 to 6 hours mean $\Delta$ECE barely moves and harm stays at 13--20 of 36 applications (per-seed ranges: 15--18 at 1 hour, 16--20 at 3, 13--17 at 6); at 12 hours both calibrators improve sharply and harm falls to 3--5 of 36, the fewest in all six seed$\times$origin combinations (by mean $\Delta$ECE, 24 hours is marginally better in one of the six), and at 24 hours it rebounds to 9--11. LR, the seed-stable model, shows the same step in every seed: 8 harmful applications of 12 at 6 hours in each seed, 1--2 at 12 hours (Table~\ref{tab:sweep}). Read as a frontier, the sweep does not price impressions: harm is common at 6 hours (median 0.74M impressions) and rare at 12 (1.86M), but the size-matched control of Section~\ref{sec:rq3} shows that 93K impressions spread over the same 12 hours do nearly as well, with a fit 20$\times$ (Platt) to 25$\times$ (isotonic) cheaper in the same runs (Table~\ref{tab:cost}). At 12 hours, isotonic recalibration lowers mean ECE on these cells from 0.0076 to 0.0043 at origin 1 and from 0.0090 to 0.0058 at origin 2 (seed 42). The two origins share slices and most test days, so the three-seed replication is a consistency check across training runs, not independent validation across time. The gap-0 cells, fitted on the night hours of the same day, harm far more at short lengths (seed 42: 10, 9, 3, 3 of 12 at 1/3/6/12 hours); pooling them with the rest (harm 25, 25, 16, 7 of 48) makes the 1-to-6-hour range look like a dose--response the fixed cells do not show.

The step is not uniform. In seed 42, five applications harm at every length tested: origin 1's DCN-v2 at gap $+2$ (isotonic) and LR at gap $+3$ (Platt; $+0.0072$, $+0.0060$, $+0.0058$, $+0.0001$, $+0.0022$ at 1/3/6/12/24 h), origin 2's LR at gap $+2$ (Platt; $+0.0057$, $+0.0067$, $+0.0057$, $+0.0015$, $+0.0028$), and origin 2's deep models at gap 0 (isotonic, 1--12 h); origin 2's LR gap-$+2$ Platt cell is positive at every length in all three seeds. The gap-$+1$ isotonic gains at origin 1 hold from 3 hours up (DCN-v2 $-0.0058$ to $-0.0064$; DeepFM $-0.0037$ to $-0.0063$) but are smaller at 1 hour ($-0.0027$, $-0.0025$); at origin 2 they are likewise smaller, and DeepFM's turns into harm at 1 hour ($+0.0017$).

The size-matched control shows the short-slice end is not a size effect: at the 1-hour impression count, spreading the impressions over 12 hours recovers most of the 12-hour gain. At the 24-hour end the slice covers the full diurnal cycle, yet harm rebounds in every seed. An aged-window control (seed 42, fixed buckets, both origins) tests staleness: 6-hour windows aged 0, 6, 12, and 18 hours give mean $\Delta$ECE of $-0.0003$, $-0.0018$, $+0.0004$, $+0.0004$ (harms 13, 8, 19, 15 of 36), not monotone in age, and the full 24-hour slice ($-0.0019$; 9 of 36) matches the best aged window, so staleness alone does not account for the rebound. Read by hour of day, the same windows are suggestive but confounded: at a fixed 6-hour span the afternoon window (hours 12--17) recovers about three-quarters of the gain of the full 12-hour slice (whose mean $\Delta$ECE is $-0.0024$), the evening window (18--23) almost none, and the hours a 24-hour slice adds (00--11) are exactly the two windows net-harmful on their own. In this design, however, a window's age and its hours of day are the same variable, and the afternoon window is also the largest (median 1.22M impressions against 0.74M for the evening window), so age, hour composition, and size cannot be separated here. With Pan et al.'s finding that calibrators fitted on two development days can also hurt~\cite{pan2020fieldaware}, the results are consistent with which hours the fitting data covers, not only how many impressions it holds, affecting whether recalibration helps; the 12 hours before the target were best in this data. That span is not expensive to occupy either: a 12-hour refit costs a median 1.41~s (Platt) or 0.87~s (isotonic); at 1-hour size the 12-hour span costs 0.06 or 0.02~s (Table~\ref{tab:cost}).

\begin{table*}[t]
\centering
\caption{Temporal arms, full grid (top: origin 1; bottom: origin 2). AUC and log loss are for the raw model; $\Delta$LL is each variant's signed log-loss change from raw; ECE for raw, Platt, and isotonic variants, fitted per bucket on the preceding 6 hours. Seed 42.}
\label{tab:grid}
\scriptsize
\renewcommand{\arraystretch}{0.6}
\begin{tabular}{llrrrrrrrr}
\toprule
Model & Bucket & AUC & Log loss & $\Delta$LL (Platt) & $\Delta$LL (Iso.) & ECE (raw) & ECE (Platt) & ECE (isotonic) \\
\midrule
LR & 0 & 0.7320 & 0.4193 & $-0.0001$ & $-0.0002$ & 0.0092 & 0.0082 & 0.0037 \\
LR & $+1$ & 0.7384 & 0.4175 & $-0.0003$ & $+0.0001$ & 0.0118 & 0.0078 & 0.0103 \\
LR & $+2$ & 0.7497 & 0.3734 & $-0.0001$ & $+0.0002$ & 0.0084 & 0.0071 & 0.0112 \\
LR & $+3$ & 0.7352 & 0.3866 & $+0.0005$ & $+0.0005$ & 0.0054 & 0.0112 & 0.0100 \\
DeepFM & 0 & 0.7431 & 0.4135 & $-0.0000$ & $-0.0000$ & 0.0069 & 0.0057 & 0.0044 \\
DeepFM & $+1$ & 0.7521 & 0.4103 & $-0.0001$ & $-0.0002$ & 0.0092 & 0.0072 & 0.0056 \\
DeepFM & $+2$ & 0.7569 & 0.3698 & $-0.0000$ & $+0.0001$ & 0.0078 & 0.0077 & 0.0073 \\
DeepFM & $+3$ & 0.7418 & 0.3834 & $+0.0001$ & $-0.0006$ & 0.0061 & 0.0070 & 0.0064 \\
DCN-v2 & 0 & 0.7408 & 0.4148 & $-0.0000$ & $-0.0001$ & 0.0070 & 0.0061 & 0.0034 \\
DCN-v2 & $+1$ & 0.7476 & 0.4125 & $-0.0001$ & $-0.0004$ & 0.0098 & 0.0086 & 0.0038 \\
DCN-v2 & $+2$ & 0.7522 & 0.3717 & $-0.0000$ & $-0.0000$ & 0.0037 & 0.0036 & 0.0042 \\
DCN-v2 & $+3$ & 0.7392 & 0.3842 & $+0.0002$ & $-0.0004$ & 0.0058 & 0.0089 & 0.0064 \\
\midrule
LR & 0 & 0.7457 & 0.4122 & $-0.0003$ & $+0.0005$ & 0.0097 & 0.0064 & 0.0069 \\
LR & $+1$ & 0.7517 & 0.3726 & $-0.0001$ & $+0.0005$ & 0.0101 & 0.0091 & 0.0127 \\
LR & $+2$ & 0.7368 & 0.3858 & $+0.0006$ & $+0.0002$ & 0.0072 & 0.0129 & 0.0102 \\
LR & $+3$ & 0.7277 & 0.4085 & $-0.0000$ & $-0.0009$ & 0.0108 & 0.0132 & 0.0132 \\
DeepFM & 0 & 0.7562 & 0.4059 & $+0.0000$ & $+0.0002$ & 0.0040 & 0.0051 & 0.0063 \\
DeepFM & $+1$ & 0.7578 & 0.3690 & $-0.0002$ & $-0.0001$ & 0.0067 & 0.0053 & 0.0051 \\
DeepFM & $+2$ & 0.7443 & 0.3825 & $-0.0002$ & $-0.0007$ & 0.0116 & 0.0082 & 0.0083 \\
DeepFM & $+3$ & 0.7381 & 0.4024 & $-0.0003$ & $-0.0003$ & 0.0103 & 0.0075 & 0.0073 \\
DCN-v2 & 0 & 0.7558 & 0.4064 & $-0.0001$ & $+0.0002$ & 0.0044 & 0.0027 & 0.0067 \\
DCN-v2 & $+1$ & 0.7573 & 0.3692 & $-0.0002$ & $-0.0001$ & 0.0064 & 0.0041 & 0.0046 \\
DCN-v2 & $+2$ & 0.7440 & 0.3824 & $-0.0001$ & $-0.0006$ & 0.0101 & 0.0084 & 0.0078 \\
DCN-v2 & $+3$ & 0.7382 & 0.4018 & $-0.0001$ & $+0.0001$ & 0.0081 & 0.0070 & 0.0077 \\
\bottomrule
\end{tabular}
\end{table*}

\begin{table}[t]
\centering
\caption{Slice-length sweep on fixed evaluation sets (gaps $+1$ to $+3$), seeds 42/43/44: mean $\Delta$ECE ($\times 10^{-4}$; negative = recalibration helps) and, in parentheses, harmful applications out of 27 (9 cells $\times$ 3 seeds). Lower rows: harms per seed (of 36), LR harms (of 12), seed-42 gap-0 harms (of 12), median slice size.}
\label{tab:sweep}
\scriptsize
\renewcommand{\arraystretch}{0.85}
\setlength{\tabcolsep}{3pt}
\begin{tabular}{llrrrrr}
\toprule
Origin & Calibrator & 1 h & 3 h & 6 h & 12 h & 24 h \\
\midrule
1 & Isotonic & $-5$ (10) & $-13$ (11) & $-11$ (9) & $-38$ (1) & $-31$ (6) \\
1 & Platt & $-4$ (8) & $-2$ (9) & $-7$ (6) & $-17$ (4) & $-11$ (11) \\
2 & Isotonic & $+18$ (18) & $+7$ (18) & $+2$ (16) & $-30$ (1) & $-30$ (3) \\
2 & Platt & $0$ (12) & $+6$ (14) & $+3$ (13) & $-14$ (6) & $-9$ (11) \\
\midrule
\multicolumn{2}{l}{Harms/seed (of 36)} & 15/15/18 & 16/16/20 & 13/14/17 & 4/5/3 & 9/11/11 \\
\multicolumn{2}{l}{LR harms/seed (of 12)} & 7/6/7 & 8/8/8 & 8/8/8 & 2/1/1 & 5/6/6 \\
\multicolumn{2}{l}{Gap-0 harms, s.\ 42 (of 12)} & 10 & 9 & 3 & 3 & --- \\
\multicolumn{2}{l}{Median slice (M impr.)} & 0.09 & 0.28 & 0.74 & 1.86 & 3.83 \\
\bottomrule
\end{tabular}
\end{table}

\subsection{RQ4 --- Profit effects are small, sign-inconsistent, and tiny even for a test-day oracle}
\label{sec:rq4}

Table~\ref{tab:bid} reports, for every model and bucket at both origins, each recalibrated strategy's simulated profit as a percentage difference from raw, alongside the label oracle's as a multiple of raw profit. Every strategy sees the same impressions and drawn market prices, so the comparisons are paired; we quantify them with a paired block bootstrap over saved per-impression predictions (200 contiguous blocks, 10,000 resamples), marking in the table the deltas whose 95\% confidence interval excludes zero. The honest summary has two parts.

First, \emph{the effects are real but sign-inconsistent, and small}. Of the 48 recalibration profit deltas, 24 have bootstrap intervals excluding zero---but they split in both directions, tracking each cell's calibration change rather than any global benefit: pooled across both origins, profit delta correlates with ECE change at $r = -0.69$ ($-0.770$ at origin 1, $-0.783$ at origin 2, from unrounded deltas; $-0.777$ and $-0.790$ from the rounded values in Tables~\ref{tab:grid} and~\ref{tab:bid}), and the sign of the profit delta tracks the sign of the ECE change in 40 of 48 cells. The 48 deltas are not independent---each cell contributes two---and the relation is not one-to-one: at origin 2, LR gap $+3$, Platt and isotonic produce nearly the same ECE change ($+0.0024$ and $+0.0025$) yet significant profit changes of opposite sign ($-0.15\%$ and $+0.15\%$). Every signed profit delta lies between $-0.207\%$ and $+0.263\%$ of raw profit---the theoretically expected scale, since in a second-price auction profit loss is second order in bid error, so calibration changes of this size should not produce first-order profit changes.

Second, \emph{test-day recalibration splits by construction}. The binned calibration oracle, which remaps each bucket's raw scores through its own 15-bin empirical click rates, earns less than the raw model in all 24 cells (23 significantly; median $-0.37$ currency units per 1,000 impressions): collapsing each bin to one bid discards within-bin gradation. The isotonic test-day oracle---the monotone map minimizing squared error on each bucket's own labels, fitted and evaluated in-sample, so an optimistic bound---beats the raw model in all 24 cells, all significantly, but by a median of only $+0.147$ per 1,000 impressions (range $+0.049$ to $+0.409$): real but tiny headroom, unreachable without the test day's labels. The simulator does register probability quality (the label oracle earns 1.75$\times$--1.97$\times$ raw profit), but that headroom is ranking information no recalibration can create.

\begin{table*}[t]
\centering
\caption{Simulated bidding profit: recalibrated strategies as \% difference from raw; label oracle as a multiple of raw profit. $^{\ast}$ = paired-bootstrap 95\% CI excludes 0 (24 of 48 deltas). Deltas under 0.005\% in magnitude show three decimals. Left: origin 1; right: origin 2.}
\label{tab:bid}
\scriptsize
\renewcommand{\arraystretch}{0.68}
\begin{minipage}[t]{0.48\textwidth}
\centering
\begin{tabular}{llrrr}
\toprule
Model & Bucket & Iso.\ $\Delta$\% & Platt $\Delta$\% & Oracle \\
\midrule
LR & 0 & $+0.02$ & $+0.03$ & 1.81$\times$ \\
LR & $+1$ & $-0.06$$^{\ast}$ & $+0.11$$^{\ast}$ & 1.79$\times$ \\
LR & $+2$ & $-0.10$$^{\ast}$ & $+0.03$ & 1.95$\times$ \\
LR & $+3$ & $-0.08$$^{\ast}$ & $-0.10$$^{\ast}$ & 1.97$\times$ \\
DeepFM & 0 & $+0.01$ & $+0.06$$^{\ast}$ & 1.78$\times$ \\
DeepFM & $+1$ & $+0.06$$^{\ast}$ & $+0.01$ & 1.75$\times$ \\
DeepFM & $+2$ & $-0.04$ & $+0.01$ & 1.93$\times$ \\
DeepFM & $+3$ & $+0.06$ & $+0.02$ & 1.95$\times$ \\
DCN-v2 & 0 & $+0.02$ & $+0.03$ & 1.79$\times$ \\
DCN-v2 & $+1$ & $+0.12$$^{\ast}$ & $+0.03$ & 1.77$\times$ \\
DCN-v2 & $+2$ & $+0.02$ & $+0.001$ & 1.95$\times$ \\
DCN-v2 & $+3$ & $-0.04$ & $-0.08$$^{\ast}$ & 1.96$\times$ \\
\bottomrule
\end{tabular}
\end{minipage}
\hfill
\begin{minipage}[t]{0.48\textwidth}
\centering
\begin{tabular}{llrrr}
\toprule
Model & Bucket & Iso.\ $\Delta$\% & Platt $\Delta$\% & Oracle \\
\midrule
LR & 0 & $+0.08$ & $+0.07$$^{\ast}$ & 1.78$\times$ \\
LR & $+1$ & $-0.19$$^{\ast}$ & $+0.05$$^{\ast}$ & 1.94$\times$ \\
LR & $+2$ & $-0.07$ & $-0.21$$^{\ast}$ & 1.96$\times$ \\
LR & $+3$ & $+0.15$$^{\ast}$ & $-0.15$$^{\ast}$ & 1.90$\times$ \\
DeepFM & 0 & $+0.01$ & $+0.002$ & 1.75$\times$ \\
DeepFM & $+1$ & $+0.04$ & $+0.06$ & 1.92$\times$ \\
DeepFM & $+2$ & $+0.23$$^{\ast}$ & $+0.26$$^{\ast}$ & 1.94$\times$ \\
DeepFM & $+3$ & $+0.21$$^{\ast}$ & $+0.14$$^{\ast}$ & 1.88$\times$ \\
DCN-v2 & 0 & $-0.11$$^{\ast}$ & $+0.004$ & 1.75$\times$ \\
DCN-v2 & $+1$ & $+0.03$ & $+0.09$$^{\ast}$ & 1.92$\times$ \\
DCN-v2 & $+2$ & $+0.23$$^{\ast}$ & $+0.19$$^{\ast}$ & 1.94$\times$ \\
DCN-v2 & $+3$ & $+0.08$ & $+0.06$$^{\ast}$ & 1.88$\times$ \\
\bottomrule
\end{tabular}
\end{minipage}
\end{table*}

\section{Discussion and Threats to Validity}
\label{sec:threats}

\textbf{Implications for evaluation practice.} The RQ1--RQ2 results argue for a concrete change in offline certification: temporal splits over several consecutive future days, at more than one training origin, with ECE reported next to AUC. A random split did not merely add noise: at origin 1's nearest test day it understated LR's ECE by 38--45\% in every seed (the deep models' by $-4\%$ to $45\%$, depending on the seed), and a single temporal holdout does not suffice either---the two origins disagree about which future day is hard, even for the seed-stable LR. A random-split calibration gate (``ship if ECE $< x$'') thus evaluates a different quantity than the serving stack will experience.

\textbf{Implications for recalibration pipelines.} A scheduled recalibration step---refit hourly, daily, or on every model refresh---applied unconditionally has a silent failure mode (RQ3): it can make the served probabilities worse---in a third of the 48 applications measured, by up to $+107\%$ in the worst cell---often when the fitting slice's mean bias misstates the incoming day's (Section~\ref{sec:rq3}; clearly so in two of three seeds)---knowable only after the day is served. On fixed evaluation sets (gaps $+1$ to $+3$), harm falls from 13--17 to 3--5 of 36 applications at 12 hours in every seed and rises again at 24 hours, though five applications harm at every length in seed 42; no tested gate removes the risk. We therefore recommend (i) treating the span and amount of fresh data a recalibration step uses as parameters to measure, not defaults---here a 12-hour span sampled sparsely came close to the full 12-hour slice at about a twentieth of its fitting cost; (ii) evaluating any gate prospectively across origins, since in-sample gate evaluations, including our own, are not evidence; and (iii) research on stronger pre-serving signals and on choosing \emph{which} calibrator to apply, since the best fixed choice flipped between origins.

\textbf{Threats to validity.} (1) \emph{Single dataset.} All measurements come from Avazu, ten days of 2014 mobile-ad traffic; which days are hard will differ on other logs. Our claims are only that per-model monotone drift is not a law and that the slice-ECE gate shows no benefit here. (2) \emph{Two overlapping origins.} The origins share most of their days, so they are two views of one ten-day span, not independent replications. Non-overlapping origins are impossible under this protocol on a ten-day log, since each origin needs nine days; a longer log or a second dataset is the necessary extension. (3) \emph{Seed variance.} The deep models carry raw-ECE seed standard deviations up to 0.0031 at origin 1 and 0.0051 at origin 2, comparable to many cell-level effects; all arms are replicated over three seeds, and we rest conclusions on seed-stable cells and pooled rates. The deep models' day-to-day ECE pattern (RQ2), their gap-0 random/temporal ECE ratio (RQ1), and the split of harms between origins (RQ3) do not reproduce across seeds. The bootstrap and RQ4 use seed 42 only. (4) \emph{Training cap.} At most 4,000,000 rows per arm; stride sampling leaves 3.34M training rows at origin 1, 3.95M at origin 2, and 4.00M in the random arm, so the arms share the cap but not exactly the training size (Section~\ref{sec:rq1}); absolute levels would shift uncapped. (5) \emph{Recalibration window.} The main protocol fits calibrators on one slice length (6 hours), and the sweep of Section~\ref{sec:sweep} shows the choice matters: harm drops sharply at 12 hours, with five persistent exceptions in seed 42; the gap-0 evaluation set changes with slice length; and an aged-window control (seed 42) finds window age alone does not explain the 24-hour rebound (age is confounded by hour of day). The gate analysis, however, is conducted only at 6 hours, and a longer slice would change the gate's input signal. (6) \emph{ECE operationalization.} ECE uses 15 equal-width bins throughout; absolute levels are not comparable across binnings. (7) \emph{Simulated auctions.} RQ4 fixes lognormal prices ($\sigma = 0.9$) and value per click 1.0, unswept; it concerns auctions with dispersed prices and raw ECE up to 0.0118, not live marketplaces. Whether calibration repair pays on surfaces that consume probabilities directly (pacing, budget forecasting, reserve prices) is unanswered by this design. (8) \emph{Bucket structure and uncertainty.} The random and temporal test sets differ in dates, hours, base rate, and size; a same-target-day comparison would isolate the protocol effect. Gap 0 differs structurally from gaps $\geq 1$: evaluated on hours 6--23 only and recalibrated on a same-day night slice rather than the preceding evening, confounding its RQ2/RQ3 comparisons. With ten days, weekday-absence coincides with gap $\in\{0,+1\}$, so the two cannot be separated. Paired block-bootstrap intervals on $\Delta$ECE (Section~\ref{sec:rq3}) hold the trained model and calibrator fixed, so they understate uncertainty relative to seed variation: 43 of 48 exclude zero.

\section{Conclusion}

On Avazu, with three standard CTR models, two rolling origins and three seeds per arm, the random split dominating offline evaluation understates calibration error in 20--23 of 24 temporal cells, by an amount that depends on the comparison day and, for the deep models, on the seed. No model's calibration error grows monotonically with distance, and the same calendar day is calibrated differently by models trained one day apart: for the seed-stable LR, distance alone does not describe the measured trajectories. With a 6-hour fitting slice, recalibration removes up to 61\% of ECE where miscalibration is large and adds error in about a third of applications; on fixed evaluation sets that harm falls by about two-thirds at a 12-hour slice and rises again at 24 hours, a pattern consistent across two overlapping origins and all three seeds, and a sample of the 12-hour slice cut to 1-hour size does nearly as well, so impression count alone is not what the fitting data must provide; which hours it covers may also matter---an afternoon 6-hour window recovered most of the 12-hour gain (seed 42), though window age, hour of day, and size are confounded there. A gate on the slice's own ECE shows no benefit on the second origin; bias-aware gates do marginally better, within noise. In the auction, calibration repair at these magnitudes moves profit by $-0.21\%$ to $+0.26\%$, and even an isotonic oracle fitted on the test day gains only a median $+0.15$ per 1,000 impressions. Measure calibration temporally, over multiple origins and seeds; choose the span and amount of fresh data a recalibration step uses by measuring them, not by default; and treat finding a pre-serving signal that anticipates tomorrow's bias as open.

\end{document}